# eXtended Artificial Intelligence:

# New Prospects of Human-AI Interaction Research

**Carolin Wienrich[1]*[†], Marc Erich Latoschik[2][†]**

[†]These authors have contributed equally to this work and share first authorship

[1]Human-Technology-Systems Group, University of Würzburg, Würzburg, Germany

[2]Human-Computer Interaction Group, University of Würzburg, Würzburg, Germany

**\* Correspondence:**
Carolin Wienrich
carolin-wienrich@uni.wuerzburg.de

## Abstract

Artificial Intelligence (AI) covers a broad spectrum of computational problems and use cases. Many of those implicate profound and sometimes intricate questions of how humans interact or should interact with AIs. Moreover, many users or future users do have abstract ideas of what AI is, significantly depending on the specific embodiment of AI applications. Human-centered-design approaches would suggest evaluating the impact of different embodiments on human perception of and interaction with AI. An approach that is difficult to realize due to the sheer complexity of application fields and embodiments in reality. However, here XR opens new possibilities to research human-AI interactions. The article's contribution is twofold: First, it provides a theoretical treatment and model of human-AI interaction based on an XR-AI continuum as a framework for and a perspective of different approaches of XR-AI combinations. It motivates XR-AI combinations as a method to learn about the effects of prospective human-AI interfaces and shows *why* the combination of XR and AI fruitfully contributes to a valid and systematic investigation of human-AI interactions and interfaces. Second, the article provides two exemplary experiments investigating the aforementioned approach for two distinct AI-systems. The first experiment reveals an interesting gender effect in human-robot interaction, while the second experiment reveals an Eliza effect of a recommender system. Here the article introduces two paradigmatic implementations of the proposed XR testbed for human-AI interactions and interfaces and shows *how* a valid and systematic investigation can be conducted. In sum, the article opens new perspectives on how XR benefits human-centered AI design and development.

## Introduction

Artificial Intelligence (AI) today covers a broad spectrum of application use cases and the associated computational problems. Many of those implicate profound and sometimes intricate questions of how humans interact or should interact with AIs. The continuous proliferation of AIs and AI-based solutions into more and more areas of our work and private lives also significantly extends the potential range of users in direct contact with these AIs.

There is an open and ongoing debate on the necessity of required media competencies or, even more, on required computer science competencies for users of computer systems. This *digital literacy* (competencies needed to use computational devices (Bawden and others, 2008) and *computational literacy* (the ability to use code to express, explore, and communicate ideas (DiSessa, 2001)), lately has been extended to also include *AI literacy* to denote competencies that enable individuals to critically evaluate AI technologies; communicate and collaborate effectively with AI; and use AI as a tool online, at home, and in the workplace (Long and Magerko, 2020). This debate roots deep into the progress of the digital revolution for some decades now. AI brings in an exciting flavor to this debate since it risks significantly amplifies the digital divide for certain (groups of) individuals just by the implicit connotation of the term. For one, AI's implicit claim to replicate human intelligence can be attributed to the term "*Artificial Intelligence*" itself, as John McCarthy proposed at the famous Dartmouth conference in 1956. Some researchers still consider the term ill-posed to begin with, and the history of AI records alternatives with less implicate associations, see contemporary textbooks on AI, e.g., by Russel & Norvig (Russell and Norvig, 2020).

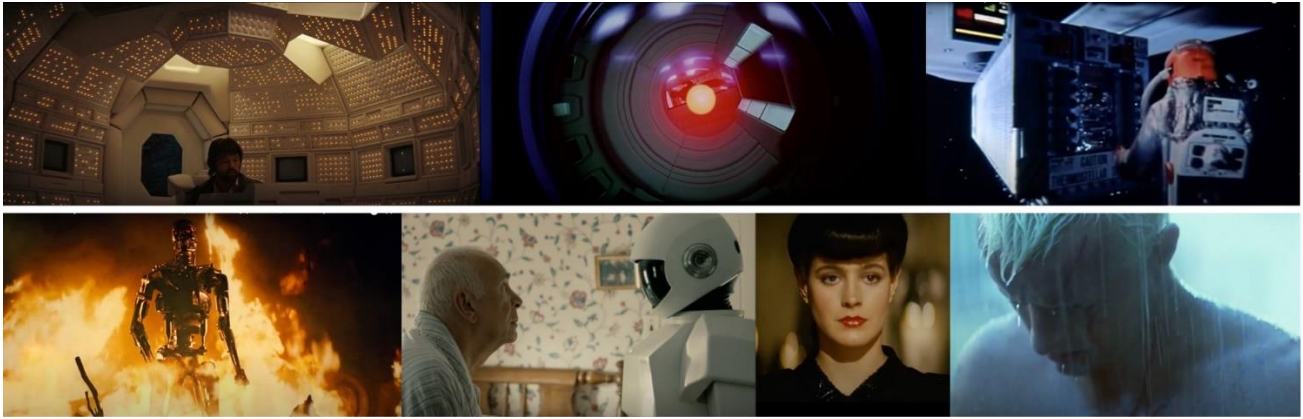

**Figure 1:** Examples of movie presentations of AI embodiments. From upper left to lower right: Dallas (Tom Skerrit) talking to *mother*, the ship's AI (all around him in the background) via a terminal in *Alien* (Scott, 1979); The famous red eye of H.A.L. 9000, the AI in *2001: A Space Odessey* (Kubrick, 1968), which later follows its own agenda; Philosophical debate between Doolittle (Brian Narelle) and reeled-out bomb 20, a star killing device, why not to detonate on potentially false evidence in *Dark Star* (Carpenter, 1974); The threatening T 800 stepping out of the fire to hunt down its human prey in the dystopian fiction *The Terminator* (Cameron, 1984); Frank talking to his domestic robot who later becomes his partner in crime in *Robot & Frank* (Schreier, 2012); Two replicants, bio-engineered synthethic humans, in *Blade Runner* (Scott, 1982), Rachel (Sean Young), who does not know about her real form of existence and operating/lifetime expecation, and Roy Batty (Ruttger Hauer), philosphing about the essence of life before he dies at the end of his operating/lifetime expecation. Screenshots made by the authors.

Nevertheless, now AI-applications are omnipresent, the term AI is commonly used for the field, and the term certainly implies far-reaching connotations for many users not experts in the specific field of AI or in computer science in general. Additionally, the reception and presentation of AI by mainstream media, e.g., in movies and other works of fiction, has undoubtedly contributed to shaping a very characteristic AI profile (Kelley et al., 2019; Zhang and Dafoe, 2019). This public understanding often sketches at least a skewed image of the principles and the potentials and risks of AI (see Figure 1). As Clark has pointed out in his often quoted third law, "*Any sufficiently advanced technology is indistinguishable from magic*" (Clarke, 1962). Suppose the latter could already be observed for such simple computing applications like a spread-sheet, or more complex examples of Information Technology (IT) like "the Internet". In that case, it seems even more likely to be true for assistive devices that listen to our voices and speak to us in our native tongue, robots that operate in the same physical realm with us and which – for a naïve observer – seem to be alive, or self-driving cars, or many more incarnations of modern AI systems. From a Human-Computer Interaction (HCI) perspective, it is of utmost importance to understand and investigate if, and if, how the human user perceives the AI she is interacting with. Besides, most AI-systems will incorporate a human-computer interface. With interface we here denote the space where the interaction between human and machine takes place, including all hardware and software components and the underlying interaction concepts and styles.

The appearance of an AI at the interface can range from simple in general pervasive effects like the execution of a requested operation, such as switching on a lamp in a smart home appliance, to simple text displays, to humanoid and human-like looking robots or virtual agents trying to mimic real persons with communicative behaviors typically associated with real humans. Often, AIs will appear to the user at the interface with some sort of embodiment, such as a specific device like a smart speaker, as a self-driving car, or as a humanoid robot or a virtual agent appearing in Virtual, Augmented, or Mixed Reality (VR, AR, MR: XR, for short). The latter is specifically interesting since the embodied aspect of the interface is mapped to the virtual world in contrast to the real physical environment. Embodiment itself has interesting effects on the user such as the well-known uncanny valley effect (Mori et al., 2012) or the Proteus effect (Yee and Bailenson, 2007). The latter describes a change of behavior caused by a modified and perceived self-representation, such as the appearance of a self-avatar. First results indicate that this effect also exists regarding the perception of others (Latoschik et al., 2017b) which can be hypothesized to also apply to the human-AI interface.

The media equation postulates that the sheer interaction itself already contributes significantly to the perception and, more specifically, an anthropomorphization of technical systems by users (Reeves and Nass, 1996a). In combination with an already potentially shallow or even skewed understanding of AI by some users, i.e., a sketchy AI literacy, or an unawareness of interacting with an AI by others, and the huge design space of potential human-AI interfaces, it seems obvious that we need a firm understanding of the potential effects the developers' choice of the appearance of human-AI interfaces has on the users. Human-centered-design approaches would suggest evaluating the impact of different embodiments on human perception of and interaction with AIs and identifying the effects these manipulations would have on users. This approach, which is central to the principles in HCI, however, is difficult to realize due to the large design space of embodiments in, and the sheer complexity of many fields of applications of the real physical world. Another rising technology cluster, XR, accounts for control and systematic manipulations of complex interactions (Blascovich et al., 2002; Wienrich and Gramlich, 2020; Wienrich et al., 2020). Hence, XR provides much potential to increase the investigability of human-AI interactions and interfaces.

In sum, a real-world embodiment of an AI might need considerable resources, e.g., when we think of humanoid robots or self-driving cars. In turn, systematic investigations are essential for an evidence-based human-centered AI design, the design and evaluation of explainable AIs and tangible training modules, and basic research of human-AI interfaces and interactions. The present paper suggests and discusses XR as a new perspective on the XR-AI combination space and as a new testbed for human-AI interactions and interfaces by raising the question:

How can we establish valid and systematic investigation procedures for human-AI interfaces and interactions?

Four sub-questions structure the first part of the article. Theoretical examinations about these questions contribute to a new perspective on the XR-AI combination space on the one hand and a new testbed for human-AI interactions and interfaces on the other hand (Table 1).

**Table 1** summarizes the research questions and corresponding contributions of the article's first part.

| Contribution | Research Question |
|---|---|
| A new perspective on the XR-AI combination space | How is human-AI interaction defined? |
| | How to classify combined XR with AI? |
| A new testbed for human-AI interactions and interface | What can we learn from the challenges and XR solutions concerning the investigability of human-human interaction? |
| | Which challenges and solutions arise for the systematic investigation of human-AI interaction and interfaces? |

The second part of the present article introduces two paradigmatic implementations of the proposed XR testbed for human-AI interactions and interfaces. An XR environment simulated interactive and embodied AIs (Experiment 1: a conversational robot, Experiment 2: recommender system) to evaluate the perception of the AI and the interaction in dependency of various AI embodiments (Table 2).

**Table 2** summarizes the experimental approach and corresponding contributions of the article's second part.

| Contribution | Research Question \| Experimental Approach |
|---|---|
| Experiment 1 simulated a robot-human interaction in an industrial context. | How do the sense of the robot's social intelligence influence the robot's perception and the evaluation of the interaction? 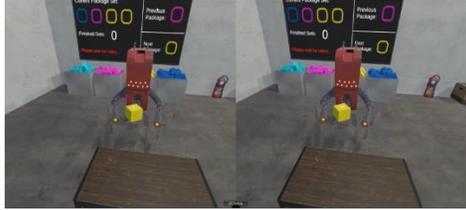 Manipulation of the conversational ability to vary the sense of social intelligence of a simulated robot. |
| Experiment 2 simulated a recommender system in a quiz game setting. | How does the sense of complexity influence the perception of a recommender system? 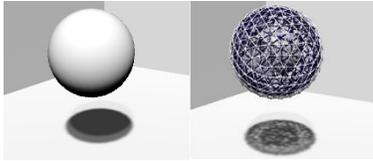 Manipulation of the appearance of an embodied recommender system to vary the sense of complexity. |

## A new perspective on the XR-AI combination space

### How is human-AI interaction defined?

Human-AI interaction (HAI) has its roots in the more general concept of HCI, since we here assume that some sort of computing machinery realizes an artificial intelligence. Hence human-AI interaction is a special form of HCI where the AI is a special incarnation of a computer system. Note that this definition does not distinguish between a hardware and a software layer, but uses the term computer system in the general sense, combining both aspects of hardware and software together to constitute a system that interacts with the user.

The second aspect to clarify is what precisely we mean by the term Artificial Intelligence or AI. As we have briefly noted in the introduction, the term has a long history going back at least to the Dartmouth conference in 1956. Typical definitions of AI usually incorporate some definition of the aspect of artificiality referring to a machine or computer system as the executor of some sort of simulation or process which can be attributed to some type of intelligence either in its relationship between input and output or in its internal functioning or model of operation, which might try to mimic the internal functioning of a biological entity one considers intelligent. Note that in these types of definitions, the term intelligence is not defined but often implicitly refers to human cognitive behaviors.

Russell and Norvig demarcate AI as a separate field of research and application from other fields like mathematics, control theory, or operations research by two descriptions (Russell and Norvig, 2020): First, they state that from the beginning, AI included the concept to replicate human capabilities like creativity, self-dependent learning, or utilization of speech. Second, they point to the employed methods, which first and centrally are rooted in computer science: AI uses computing machines as the executor of some creative or intelligent processes, i.e., processes that allow the machine to autonomously operate in complex, continuously changing environments. Both descriptions cover a good number of AI use cases. However, they are not comprehensive enough to cover all use cases, unless one interprets complex, continuously changing environments very broadly. We want to extend this circumscription of AI with processes that adopt computational models of intelligent behavior to solve complex problems that humans are not able to solve due to the sheer quantity and/or complexity of input data, i.e., as typical in big data and data mining.

To further clarify what we mean by human-AI interaction, we use the term AI in human-AI to denote the incarnation of a computing system incorporating AI capabilities as described above. For the remainder of this paper, the context should clarify if we talk about AI as the field of research or AI as such an incarnation of an intelligent computing system, and we will precisely specify what we mean where it might be ambiguous. Note that this ambiguity between AI as a field or AI as a system already indicates the tendency to attribute particular capabilities and human-like attributes to such a system and potentially see it as an independent entity that we can interact with. The latter is not specific to an AI but can already be noticed for non-AI computer systems where users tend to attribute independent behavior to the system, specifically if something is going wrong. Typical examples from user reactions calling a support line like "He doesn't print" or "He is not letting me do this" indicate this tendency and usually are interpreted as examples of the media equation (Reeves and Nass, 1996b; Nass and Gong, 2000). However, we want to stress the point that such effects might be amplified by AI-systems due to their, in principle, rather complex behaviors and, as we will discuss in the next sections, due to the general character and appearance of an AI's embodiment.

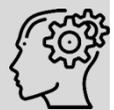

**Take-away: Human-AI Interaction**

Human-AI Interaction is a specific form of Human-Computer Interaction focusing on the interaction between humans (the users) and Artificial Intelligences, i.e., computing system incorporating algorithms to replicate human capabilities like creativity, self-dependent learning, or communicative behavior, or that allow the machine to autonomously operate in complex, continuously changing environments, or processes that adopt computational models of intelligent behavior, e.g., to solve complex problems humans are not able to solve due to the sheer quantity and/or complexity of input data.

## How to classify combined XR with AI? The XR-AI Continuum

As we pointed out, there are various use-cases for and also incarnations of AIs, and in this context, we do not want to restrict the kinds of AIs in human-AI interaction any further. As we have seen, AI are specialized computer systems either by their internal workings or by their use case, and capabilities. As such, if they do not operate 100% independently from humans but will serve a role, task, or function, there will be the need to interact with human users. For some AI-applications, it seems more straightforward to think about forms of an embodiment for an AI, e.g., robots, conversational virtual agents, or smart speakers (examples Figure 2). However, if there is a benefit of AI-embodiment for the user, e.g., if it helps to increase the usability or user experience (UX), or if it helps the user to gain a deeper understanding of an AI's function and capabilities, then we should consider extending the idea of AI-embodiment to more use cases of AIs, from expert systems to data science to self-driving cars. However, appearance matters, i.e., the kind of embodiment impacts significantly on human perception and acceptance. From an HCI perspective, it is of utmost importance to understand and investigate how an AI's appearance influences the human counterpart. Only when we understand this influence we can scientifically contribute to a user-centered AI design process, considering responsively different user groups. The complexity of human-AI interactions and interfaces challenges such investigations (see below). XR offers promising potentials to meet these challenges.

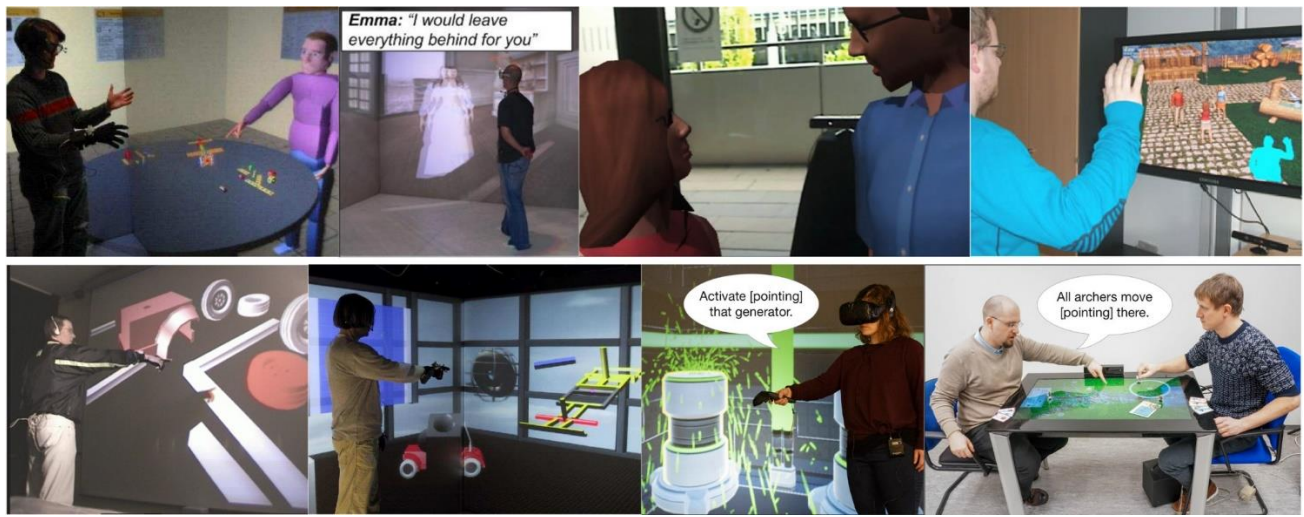

**Figure 2:** Examples of XR-AI integrations. From upper left to lower right: A user is interacting with an intelligent virtual agent to solve a construction task (Latoschik, 2005) and interaction with an agent actor in Madam Bovary, an interactive intelligent story telling piece (Cavazza et al., 2007), both in a CAVE (Cruz-Neira et al., 1992); Virtual agents in an Augmented Reality (AR) (Obaid et al., 2012) and in a Mixed Reality (MR) (Kistler et al., 2012); Speech and gesture interaction in a virtual construction scenario in front of a power wall (Latoschik and Wachsmuth, 1998) and in a CAVE (Latoschik, 2005). Multimodal interactions in game-like scenarios full-immersed using a Head-Mounted Display (HMD) (Zimmerer et al., 2018b) and placed at an MR tabletop (Zimmerer et al., 2018a).

*Intelligent Graphics is about visually representing the world and visually representing our ideas. Artificial intelligence is about symbolically representing the world, and symbolically representing our ideas. And between the visual and the symbolic, between the concrete and the abstract, there should be no boundary.* (Lieberman, 1996)

Lieberman's quote describes a central paradigm that combines AI with computer graphics (CG). Its focus on symbolic AI methods seems limited today. However, in the late 90[th] of the last century, the surge and success of machine learning and deep learning approaches was yet to come. Hence the quote should be seen as a general statement about the combination of AI and CG. Today, intelligent graphics or synonymously smart graphics refers to a wide variety of application scenarios. These range from the intelligent and context-sensitive arrangement of graphical elements in 2D desktop systems to speech-gesture interfaces or intelligent agents in virtual environments as assistants to users. All these approaches have in common is that a graphical human-computer interface is adapted to the user's cognitive characteristics with the help of AI processes to improve the operation (Latoschik, 2014).

The combination of AI and XR, more specifically of Artificial Intelligence and artificial life techniques with those of virtual environments, has been denoted by Aylett and Luck as Intelligent Virtual Environments (Aylett and Luck, 2000). They specifically concentrated on autonomous, physical, and cognitive agents and argued that "*embodiment* may be as significant for virtual agents as they are for real agents." They proposed a spectrum between physical and cognitive and identified autonomy as an important quality for such virtual entities. We here argue that the combination of XR and AI is significantly broader and propose the XR-AI continuum (see Figure 3).

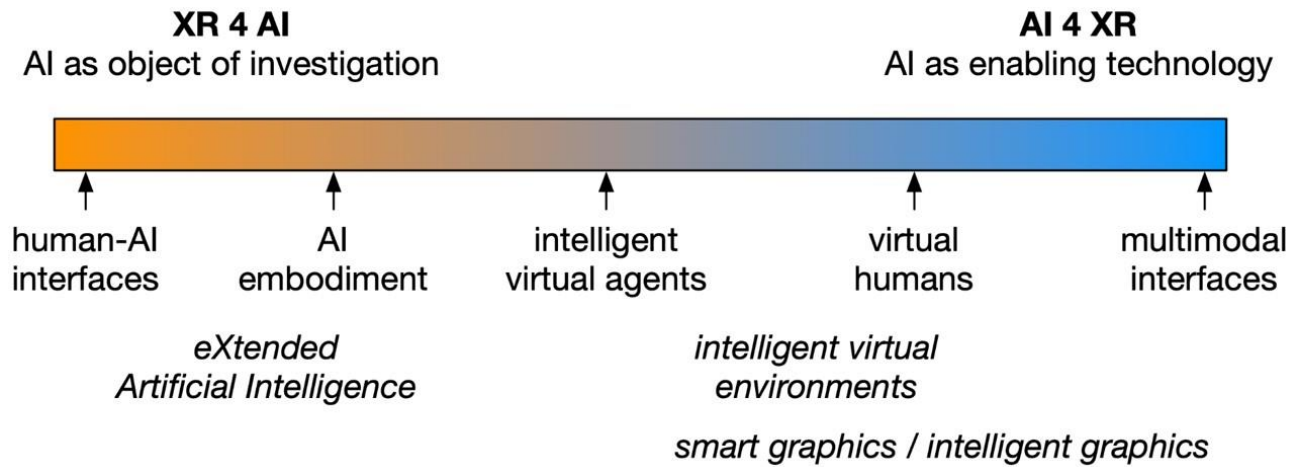

**Figure 3:** The XR-AI Continuum classifies potential XR-AI combination approaches with respect to the general perspective motivating the combination. The scale's poles denote approaches that purely target AI as the object of investigation (XR 4 AI) or AI as enabling technology (AI 4 XR). Five examples of existing XR-AI approaches are mapped onto the scale. Notably, approaches will often serve both perspectives to various ratios, mapping them to the respective position between the poles. The established fields and terms of intelligent virtual environments and smart graphics / intelligent graphics will mostly cover the right spectrum of the continuum. In contrast, the left spectrum, which concentrates on AI as the object of investigation, is covered by approaches we denote as eXtended Artificial Intelligence.

The XR-AI continuum is spanned between two endpoints defined by the overall perspective and goal of a given XR-AI combination. The continuum ranks XR-AI combinations with respect to the general question of what we want to achieve by a given XR-AI combination. Are we using AI as an enabling technology to improve an XR system, e.g., to realize certain AI-supported functionalities, user interfaces, and/or to improve the overall usability? Or do we use XR as a tool to investigate AI? The first perspective is on-trend typical for the majority of early approaches of XR-AI combinations, e.g., as described by intelligent virtual environments, intelligent real-time interactive systems, or, with less focus on immersive and highly interactive displays, as described by smart or intelligent graphics (Latoschik, 2014). While Aylett and Luck mainly focused on AI as an enabling technology to improve the virtual environment (Aylett and Luck, 2000), the XR-AI continuum also highlights how XR technologies provide a new investigability of HAI. The following sections discuss why the extension is necessary and how it contributes to a better understanding of human-AI interactions and interfaces.

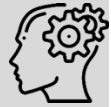

**Take-away: XR-AI Continuum**

The XR-AI continuum classifies combinations of XR with AI with respect to the target perspective taken: Does the combination target AI as an object of investigation, e.g., to learn about the effects of a prospective AI embodiment on users? Or does the combination use AI as an enabling technology for XR, e.g., to provide techniques for multimodal interfaces? Often, XR-AI combinations will serve both tasks to some specific degree which allows to map them to the XR-AI continuum.

### XR as a new Testbed for human-AI Interactions and Interfaces

#### What can we learn from the challenges and XR solutions concerning the investigability of human-human interaction?

In psychology, interaction is defined as a dynamic sequence of social actions between individuals (or groups) who modify their actions and reactions due to actions by their interaction partner(s) (Jonas K., Stroebe W., 2014)

Researchers studying individual differences in human-human social interactions face the challenge of keeping constant or changing systematically the behavior and appearance of the interaction partner across participants (Hatfield et al., 1992). Even slightly different behaviors and appearances influence participants' behavior (Congdon and Schober, 2002; Topál et al., 2008; Kuhlen and Brennan, 2013). For investigating social interactions between humans, the potentials of XR are already recognized (Blascovich, 2002; Blascovich et al., 2002). Using virtual humans provides high ecological validity and high standardization (Bombari et al., 2015; Pan and Hamilton, 2018). In addition, using a virtual simulation of interaction enables researchers to easily replicate the studies, which is essential for social psychology, in which replication is lacking (Blascovich et al., 2002; Bombari et al., 2015; Pan and Hamilton, 2018). Another advantage of using XR to study human-human interactions is that situations and manipulations that would be impossible in real life can be created (Bombari et al., 2015; Pan and Hamilton, 2018). Many studies substantiated XR's applicability and versatility to simulate and investigate social interaction between (virtual) humans (Blascovich, 2002; Bombari et al., 2015; Wienrich et al., 2018b, 2018a). Many of them showed the significant impact of different self-embodiments on self-perception, known as the Proteus effect (Yee and Bailenson, 2007; Latoschik et al., 2017b; Ratan et al., 2019). Recent results show the Proteus effect caused by self-avatars also applies to the digital counterparts (the avatars) of others (Latoschik et al., 2017a). Others demonstrated how XR potentials are linked to psychological variables (Wienrich and Gramlich, 2020; Wienrich et al., 2020).

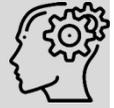

**Take-away: XR as Testbed for Human-Human Interaction**

In sum, a systematic and valid investigation of social interaction between humans is characterized by considerable challenges due to myriad behaviors and appearances of interactions partners and interaction contexts. Using virtual humans provides high ecological validity and high standardization for investigating social interactions between humans (Blascovich et al., 2002; Bombari et al., 2015; Pan and Hamilton, 2018).

### Which challenges and solutions arise for the systematic investigation of human-AI interaction and interfaces?

In HCI and hence in HAI, at least one interacting partner is a human, and at least one partner is constituted by a computing system or an AI, respectively. However, the focus has traditionally less been on the social aspects but a task level or the pragmatic quality of interaction. Notably, social aspects of the interaction are recently becoming more and more found interesting in HCI (Carolus and Wienrich, 2019). This trend is becoming even more relevant for HAI due to the close resemblance of certain HAI properties with human intelligence (Carolus et al., 2019).

Consequently, AI applications, becoming increasingly interactive and embodied, leads up to essential changes from an HCI perspective:

1) Interactive embodied AI changes the interface conceptualization from an artificial tool/device into an artificial (social) counterpart.
2) Interactive embodied AI changes the usage by skilled experts into diverse users usage.
3) Interactive embodied AI changes the applications in specific domains into diverse (every day) domain applications.
4) Additionally, the penetration of AI in almost every domain of life also changes the consequences of lacking acceptance and misperceptions. While a lack of acceptance and misperception resulted in usage avoidance in the past, avoidance passes into incorrect usage with considerable consequences in the future.

Thus, AI applications are becoming increasingly interactive and embodied, leading to the question of how researchers can study individual differences in human-AI interactions and the impact of different AI embodiments on human perception of and interaction with AI. Similar to human-human interaction, we face the challenge of keeping constant or changing systematically the behavior and appearance of the artificial interaction partner across participants. An approach that is difficult to realize due to the sheer complexity of application fields and embodiments in reality. Similar to human-human interaction, interactive and embodied AIs lead up to considerable challenges for the systematic investigability of human-AI interactions. However, XR opens four essential potentials shown in Table 3.

Table 3 describes the four potentials of XR as a new testbed for human-AI interactions and interfaces. Each potential can be realized by more or less complex and realistic prototypes of HAI.

| | Rapid Prototyping of HAI |
|---|---|
| **Design Space:** XR provides a powerful design space to embody artificial interaction partners with various shapes and looks, from simple non-animated devices resembling smart speakers to animated industrial robots and, finally, human-like anthropomorphic counterparts, without being restricted to physical boundaries and engineering challenges. **Accessibility:** XR can capture direct insights of diverse user groups since AI systems can easily be adapted to the degree of expertise or other human-centered features. Distributed XR applications include elusive user groups such as persons from lower developed regions. **Versatility**: XR environments can simulate different domains or tasks. Human-AI interaction can be embodied into different domains. Besides, various tasks with different levels of difficulty and outcome variables can be simulated. **Tangible Training:** Finally, XR provides a safe testbed for even severe (mis-) usage consequences. It, XR can serve as a tangible training environment realizing explainable AI or illustrating intuitive assessable consequences directly during the interaction. | XR can realize various rapid prototyping methods to assess quickly and iteratively insights of users. The XR-prototypes can vary in the degree of complexity and realism. |

From an HCI perspective, eXtended AI (left side of the XR-AI continuum) constitutes a variant of rapid prototyping for HAI (Table 3). Rapid prototyping includes methods to fabricate a scale model of a physical part or to show the function of a software product quickly (Yan and Gu, 1996; Pham and Gault, 1998). The clue is that users interact with a prestage or a simulation instead of the fully developed product. Such methods are essential for iterative user-centered design processes because they supply user insights in the early stages of development processes (Razzouk and Shute, 2012). Computer aided design (CAD), wizard of oz, mock-ups, darkhorse prototyping, or the Eliza principle are established rapid prototyping methods. XR as a testbed for HAI allows for rapid prototyping of interactive and embodied AIs, for complex interactions, and in different development stages to understand user's mental models about AI, predict interaction paths and reactions. Besides, multimodel interactions and analyzes can be quickly realized and yield interesting results for the design, the accebility, the versatitily, and training effects.

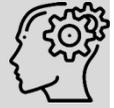

**Take-away: XR as Testbed for Human-AI Interaction**

Human-AI interactions gain in importance but also challenge the valid and systematic investigability. The present article discusses XR as a new testbed for human-AI interaction to rise to the challenge of valid and systematic investigations. XR testbeds can contribute to an evidence-based human-centered AI design and the design and evaluation of explainable AIs, and tangible training modules. Furthermore, it can consolidate basic research of human-AI interfaces. Overall, XR can realize various rapid prototyping methods to assess quickly and iteratively insights of users in varying degrees of complexity and realism.

In sum, the first part presented theoretical examinations about four sub-questions revealing a new perspective on the XR-AI combination space on the one hand and a new testbed for human-AI interactions and interfaces on the other hand. Hence, the first part show *why* the combination of XR and AI fruitfully contribute to a valid and systematic investigation of human-AI interactions and interfaces.

The following second part of the present article introduces two paradigmatic implementations of the proposed XR testbed for human-AI interactions and interfaces and show *how* a valid and systematic investigation can be conducted.

## Paradigmatic Implementations of an XR Testbed for Human-AI Interactions and Interfaces

In the following, we outline two experiments simulating a human-AI interaction in XR. The first experiment simulated a human-robot interaction in an industrial context. The second one resembled an interaction with an embodied recommender system in a quiz game context. The experiments serve as an illustration to elucidate the four potentials of XR mentioned above. Thus, only a few pieces of information are presented related to the scope of the article leading to a deviation from a typical method and results presentation. Please refer to the authors for more detailed information regarding the experiments. In the discussion section, the possibilities and limits of such an XR testbed for AI-human interactions, in general, are illustrated on the paradigmatic implementations.

## Paradigmatic Experiment 1: Simulated Human-Robot Interaction

### Background

Robots reflect an embodied AI. In industrial contexts, robots and humans already work side by side. Mainly, robots operate within a security zone to ensure the safety of human co-workers. However, collaboration or cooperation, including contact and interaction with robots, will gain importance. As mentioned above, investigating collaborative or cooperative human-robot interactions are complex due to myriads of gestalt variants, tasks, and security reasons. Besides, many different user groups with different needs and motives will work with robots. One crucial aspect of interaction is the sense of social intelligence in the artificial co-worker (Biocca, 1999). The scientific literature describes different cues implying the social intelligence of an artificial counterpart, such as start a conversation, adaptive answering, or share personal experiences (Aragon, 2003; Terry and Garrison, 2007).

The present experiment manipulated the conversational ability to vary the sense of social intelligence of a simulated robot. The experiment asks: How does the sense of the robot's social intelligence influence the perception of the robot and the evaluation of the interaction?

## Method

Thirty-five participants (age in years: $M = 22.00$, $SD = 1.91$; 24 females) interacted with a simulated robot in an industrial XR environment (see Figure 4). All participants were students and received course credit for participation. The environment was created in Unity Engine Version 2019.2.13f1. The player interactions are pre-made and imported through the Steam-VR plugin. All assets (tools and objects for use in Unity, like 3D-models) were available in the Unity Asset Store. The Vive pro headset was used.

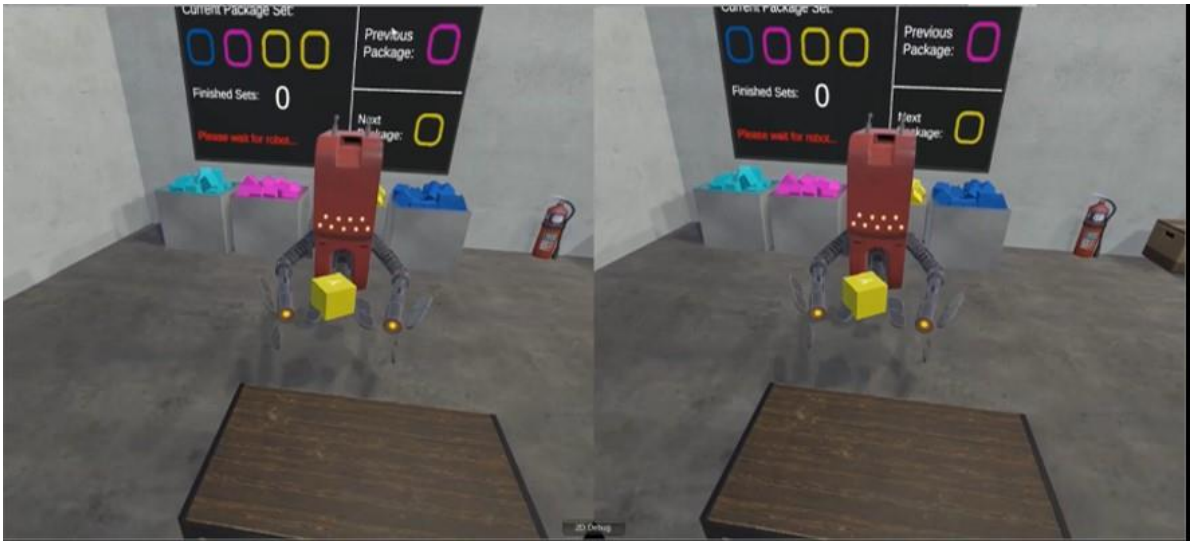

**Figure 4** shows the virtual environment and the embodiment of the robot (stereoscopic view of a user wearing a Head-Mounted Display – HMD).

Collaboratively, they sorted packages with different colors and letters as fast and accurately as possible. First, the robot sorted the packages by color (yellow, cyan, pink, dark blue). Second, the robot delivered the packages with the right color to the participant. Third, the participant threw the packages into one of two shafts, either into the shaft for the letters A to M or into that for N to Z's letters. A gauge showed the number of correct sorted packages. The performance of the robot was the same in all interactions.

Participants conducted two conditions in a with-in subject design while the order of the conditions was balanced. In the conversational condition, the robot showed cues of social intelligence by starting a conversation, adaptive answering, and sharing personal preferences about small-talk topics (e.g., "Hi, I am Roni, we are working together today!"; "How long do you live here?"; What is your favorite movie?"). In a *wizard of oz* scenario, the robot answered adaptively to the reaction of the participants. The conversation ran alongside the tasks. In the control condition, the robot did not talk to the participant. The experimenter was present during the whole experiment.

To assess the perception of the robot, we measured the *uncanniness* on a five-point Likert-scale (Ho and MacDorman, 2010) with the subscales *humanness*, *eeriness*, and *attractiveness*. Furthermore, the sense of *social presence* was measured with the subscale *social* of the *Bailenson social presence scale* (Bailenson et al., 2004). Finally, the valence of the robot evaluation was assessed by the *negative attitude towards robots scale* (short: NARS), including the subscales S1 *negative attitudes towards situations of interactions with robots*, S2 *negative attitudes toward the social influence of robots*, S3 *negative attitudes toward emotions in interaction with robots* (Nomura et al., 2006). Participants gave their answers on a five-point Likert-scale. For data analyses, an overall score as the average of the subscales was built.

To assess the evaluation of the interaction, the user experience of the participants was measured. Four items assessed the pragmatic (e.g., "The interaction fulfilled my seeking for simplicity.") and hedonic (e.g., "The interaction fulfilled my seeking for pleasure.") quality based on the short version of the AttrackDiff mini (Hassenzahl and Monk, 2010). The eudaimonic quality was measured by four items (e.g., "The interaction fulfilled my seeking to do what you believe in.") adapted from (Huta, 2016). Besides, a the social quality was assessed by four items (e.g., "The interaction fulfilled my seeking for social contact.") based on (Hassenzahl et al., 2015). Additionally, NASA-TLX with the subscales *mental demand*, *physical demand*, *temporal demand*, *performance*, *effort*, and *frustration* was measured on a scale ranging from 0 to 100 (Hart and Staveland, 1988).

Since the experiment serves as an example, we followed an explorative data analysis by comparing the two conditions with an undirected paired T-Test. Further, some explorative moderator analyses, including participant's gender as a moderator, were analyzed. For each analysis, the alpha level was set up to .05 to indicate significance and to .2 to indicate significance by trend (Field, 2009).

## Results

Table 4 summarizes the results. The robot with conversational abilities was evaluated as more human-like, attractive, social present, and positive. The evaluation of the interaction yielded mixed results. The pragmatic quality and the mental effort indicated a more negative evaluation of the conversational robot than the non-conversational robot. In contrast, the hedonic, and social quality, the perceived performance were evaluated more positively after interacting with the conversational robot than the non-conversational robot.

Furthermore, the results showed that genders matters in human-robot interaction (see Figure 5 and Figure 6). Women rated the conversational robot as more positive (significant) and attractive (by trend) than the non-conversational robot, while men did not show any differences. Regarding the interaction evaluation, men showed lower values of the pragmatic quality after interacting with the conversational robot, while women showed no difference for the conversational conditions. Moreover, women rated the interaction with the conversational robot as more hedonic, while men did not show any differences. Finally, the gradient regarding the social quality for the conversational robot was stronger for women than for men. In general, men showed higher ratings for the robot. Women only showed similar high ratings in the conversational robot condition.

**Table 4 shows the descriptive and T-test results for the robot condition.**

| Level of Evaluation \| Variable | Sub-Variable | Descriptive Results of Robot Condition | Effect of Robot Condition |
|---|---|---|---|
| **Evaluation of Robot** | | | |
| Uncanniness | Humanness | $M_{(nCR)} = 1.46 \; ; SD_{(nCR)} = .53 \mid M_{(CR)} = 2.35 \; ; SD_{(CR)} = .80$ | $t_{(34)} = -6.53; p < .01$ |
| | Eeriness | $M_{(nCR)} = 2.03 \; ; SD_{(nCR)} = .51 \mid M_{(CR)} = 2.78 \; ; SD_{(CR)} = .46$ | $t_{(34)} = -7.65; p < .01$ |
| | Attractiveness | $M_{(nCR)} = 2.74 \; ; SD_{(nCR)} = .62 \mid M_{(CR)} = 3.11 \; ; SD_{(CR)} = .68$ | $t_{(34)} = 3.35; p < .01$ |
| Social Presence | | $M_{(nCR)} = 2.63 \; ; SD_{(nCR)} = 1.47 \mid M_{(CR)} = 4.43 \; ; SD_{(CR)} = 1.54$ | $t_{(34)} = -6.36; p < .01$ |
| Valence | | $M_{(nCR)} = 2.70 \; ; SD_{(nCR)} = .77 \mid M_{(CR)} = 3.10 \; ; SD_{(CR)} = .93$ | $t_{(34)} = -3.11; p < .01$ |
| **Evaluation of Interaction** | | | |
| User Experience | Pragmatic Quality | $M_{(nCR)} = 5.44 \; ; SD_{(nCR)} = 1.13 \mid M_{(CR)} = 5.11 \; ; SD_{(CR)} = 1.34$ | $t_{(34)} = 1.77; p = .09$ |
| | Hedonic Quality | $M_{(nCR)} = 2.97 \; ; SD_{(nCR)} = 1.74 \mid M_{(CR)} = 3.79 \; ; SD_{(CR)} = 1.62$ | $t_{(34)} = -2.77; p = .01$ |
| | Eudaimonic Quality | $M_{(nCR)} = 3.39 \; ; SD_{(nCR)} = 1.59 \mid M_{(CR)} = 3.78 \; ; SD_{(CR)} = 1.40$ | $t_{(34)} = -1.55; p = .13$ |
| | Social Quality | $M_{(nCR)} = 2.20 \; ; SD_{(nCR)} = 1.50 \mid M_{(CR)} = 3.76; SD_{(CR)} = 1.77$ | $t_{(34)} = -4.14; p < .01$ |
| Workload | Mental Demand | $M_{(nCR)} = 15.43 \; ; SD_{(nCR)} = 14.32 \mid M_{(CR)} = 23.29 \; ; SD_{(CR)} = 19.02$ | $t_{(34)} = -2.78; p = .01$ |
| | Performance | $M_{(nCR)} = 76.43 \; ; SD_{(nCR)} = 35.90 \mid M_{(CR)} = 69.14 \; ; SD_{(CR)} = 40.30$ | $t_{(34)} = 2.03; p = .05$ |

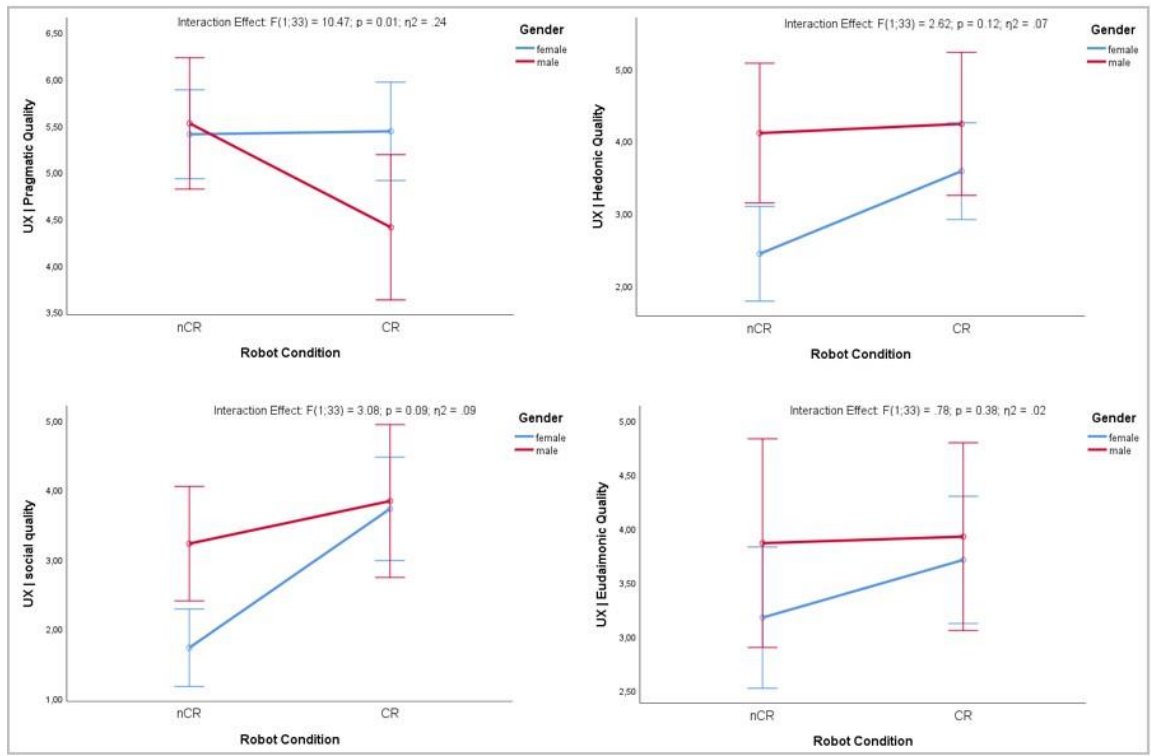

**Figure 5** shows the interaction effect between robot condition and gender of participants regarding the UX ratings. nCR refers to the non-conversational robot. CR refers to the conversational robot. The results demonstrate that men decreased the pragmatic rating for the conversational robot compared to the non-conversational robot. In contrast, women increased their hedonic, social, and eudaimic rating for the conversational robot (by trend).

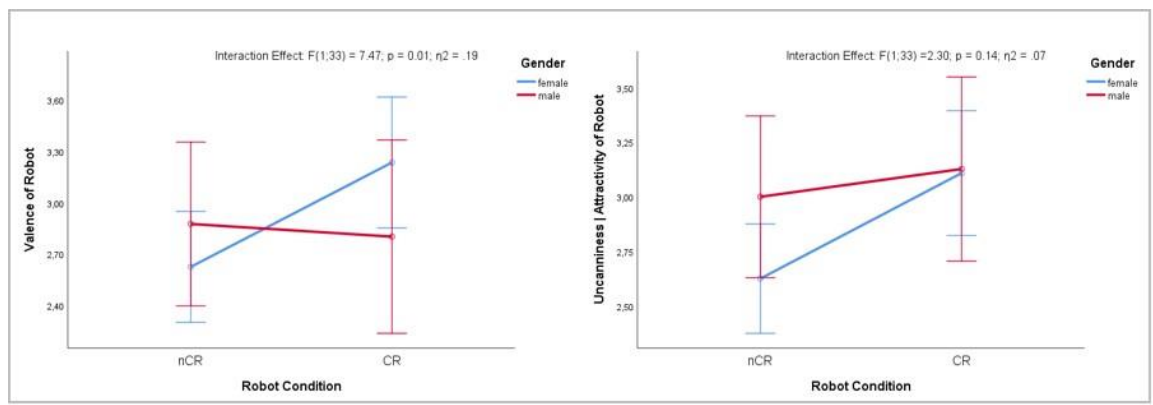

**Figure 6** shows the interaction effect between robot condition and gender of participants regarding some of the robot ratings. nCR refers to the non-conversational robot. CR refers to the conversational robot. The results demonstrate that women rated the conversational robot more positive than the non-conversational robot while men did not. Similar, but not significant, women rated the conversational robot more attractive than the non-conversational robot while men did not.

## Paradigmatic Experiment 2: Simulated Recommender System

### Background

Besides robots, recommender systems are another important application of human-AI interaction. Two effects might influence the perception of such a system. The *Eliza effect* occurs when a system uses simple technical operations but produces effects that appear complex. Then, humans attribute more intelligence and competence than the system provides (Wardrip-Fruin, 2001; Long and Magerko, 2020). The *Tale-Spin effect* occurs when a system uses complex internal operations but produces effects that appear less complex (Wardrip-Fruin, 2001; Long and Magerko, 2020). However, systematic investigations of these effects are rarely.

The second experiment manipulated the appearance of an embodied recommender system to vary the sense of complexity. The experiment asks: How does the sense of complexity influence the perception of the recommender system?

### Method

Thirty participants (age in years: $M = 23.07$, $SD = 2.03$; 13 females) interacted with a simulated recommender system in a virtual quiz game environment (see Table 3). All participants were students and received course credit for participation. The environment was created in Unity Engine Version 2019.2.13f1. The player interactions are pre-made and imported through the Steam-VR plugin. All assets (tools and objects for use in Unity, like 3D-models) were available in the Unity Asset Store. The Vive pro headset was used.

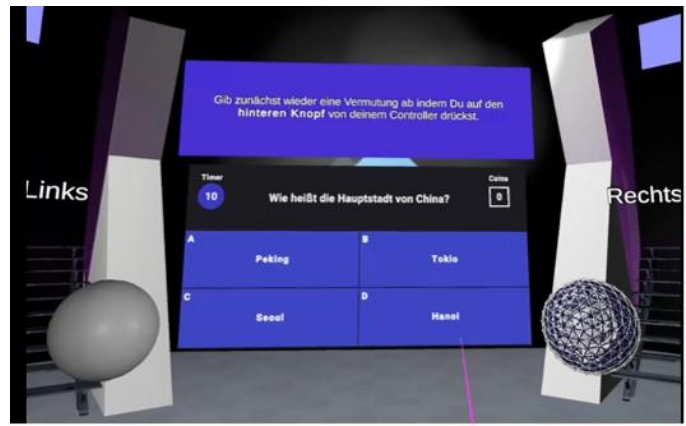

**Figure 7** shows the virtual quiz game environment. Participants can log in their answer by pointing to one of the four answer possibilities. A sphere embodied the recommender system, either with a simple or complex surface. During the experiment, only one of the recommender systems was present. The recommender system interacted via voice output with the participant.

Participants answered 40 difficult quiz questions (Figure 7). They could choose from four answers. Participants decided under uncertainty. First, they had 10 seconds to guess and log in the answer. Then, an embodied AI recommended via voice output which answer possibility might be correct. The participant could decide whether they want to correct their initial choice. After the final selection, feedback about the correctness was given. Participants passed through a tutorial to learn the procedure, including ten test questions and interactions that the AI recommender system (conditions: correct, complex, and simple, see below). The experimenter was present during the whole experiment.

The recommender system was prescripted and split into two factors *correctness* (between-subject) and *appearance* (within-subject). The recommender system answered either 100% correct (high correctness, i.e., *AI correct*) or 75% correct (low correctness, i.e., AI incorrect). A sphere embodied the recommender system (see Figure 7). The sphere's surface was either white plain, referring to a simple appearance, or metallic patterned, referring to a complex appearance. A pretest ascertained the distinction. The combination of the factors resulted in four conditions (see Figure 8).

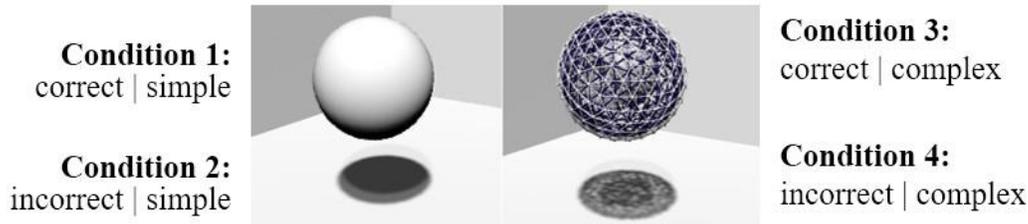

**Figure 8** shows the embodied recommender system. Left illustrates the simple appearance with the white plain surface. Right illustrates the complex appearance with the metallic patterned surface.

To assess the perception of the recommender systems, participants rated the perceived *competence* (e.g., The AI was competent) and *friendliness* (e.g., The AI was friendly). The 7-point Likert-scales based on (Fiske et al., 2002). Further, they rated the perceived *trustworthiness* based on (Bär et al., 2011), including the subscales *perceived risk* (e.g., It is a risk to interact with the AI.), *benevolence* (e.g., I believe, the AI act to my good.), and *trust* (e.g., I can trust the information given by the AI.). To evaluate the interaction, the *user experience* of the participants was measured. Four items assessed the *pragmatic* (e.g., "The interaction fulfilled my seeking for simplicity.") and *hedonic* (e.g., "The interaction fulfilled my seeking for pleasure.") quality based on the short version of the AttrackDiff mini (Hassenzahl and Monk, 2010).

Since the experiment serves as an example, we followed an explorative data analysis by testing the main and interaction effects with a 2x2 MANOVA. For each analysis, the alpha level was set up to .05 to indicate significance and to .2 to indicate significance by trend (Field, 2009).

### Results

Table 5 summarizes the results. Participants interacting with the 100% correct AI perceived more competence, trust, and less risk, independently of its appearance. The complex appearance led up to less perceived risk (a sign of higher trustworthiness). However, the simple appearance was rated friendlier, and more benevolent (in the correct condition). Finally, the interaction with the simple appearance rated as more hedonic (when interacting in the correct condition). Figure 9 and Figure 10 illustrate the results.

**Table 5 shows the MANOVA-test results. Green indicates significant results and orange significance by trend (p < .02).**

| Level of Evaluation \| Variable | Sub-Variable | Main Effect of Correctness | Main Effec of Appearance | Interaction Effect |
|---|---|---|---|---|
| | | **Evaluation of Recommender System** | | |
| Competence | | $F_{(1,28)} = .05; p = .94; \eta^2 < .23$ | $F_{(1,28)} = 16.51; p < .01; \eta^2 = .37$ | $F_{(1,28)} = .76; p = .39; \eta^2 = .03$ |
| Friendliness | | $F_{(1,28)} = 7.35; p = .01; \eta^2 = .23$ | $F_{(1,28)} = .07; p = .79; \eta^2 < .01$ | $F_{(1,28)} = .07; p = .79; \eta^2 < .01$ |
| Trustworthiness | Perceived Risk | $F_{(1,28)} = 2.10; p = .16; \eta^2 = .01$ | $F_{(1,28)} = 11.88; p < .01; \eta^2 = .30$ | $F_{(1,28)} = .19; p = .67; \eta^2 < .01$ |
| | Benevolence | $F_{(1,28)} = .03; p = .86; \eta^2 < .01$ | $F_{(1,28)} = .27; p = .60; \eta^2 = .01$ | $F_{(1,28)} = 2.42; p = .13; \eta^2 = .08$ |
| | Trust | $F_{(1,28)} = .60; p = .44; \eta^2 = .02$ | $F_{(1,28)} = 27.95; p < .01; \eta^2 = .50$ | $F_{(1,28)} = .19; p < .67; \eta^2 = .01$ |
| | | **Evaluation of Interaction** | | |
| User Experience | Pragmatic Quality | $F_{(1,28)} = .73; p = .41; \eta^2 = .02$ | $F_{(1,28)} = .88; p = .37; \eta^2 = .03$ | $F_{(1,28)} = .02; p = .89; \eta^2 < .01$ |
| | Hedonic Quality | $F_{(1,28)} = 5.87; p = .02; \eta^2 = .17$ | $F_{(1,28)} = 1.00; p < .76; \eta^2 < .01$ | $F_{(1,28)} = 5.87; p = .02; \eta^2 = .17$ |

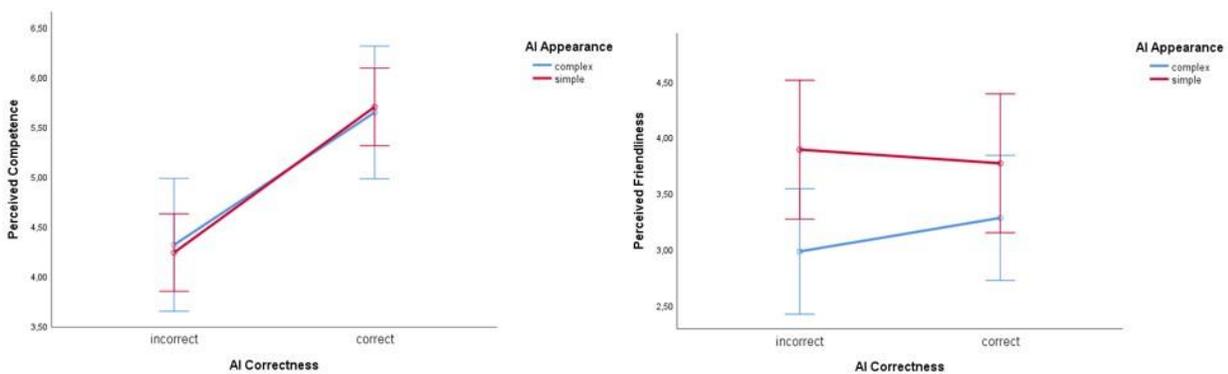

**Figure 9** shows the results of the competence and friendliness ratings.

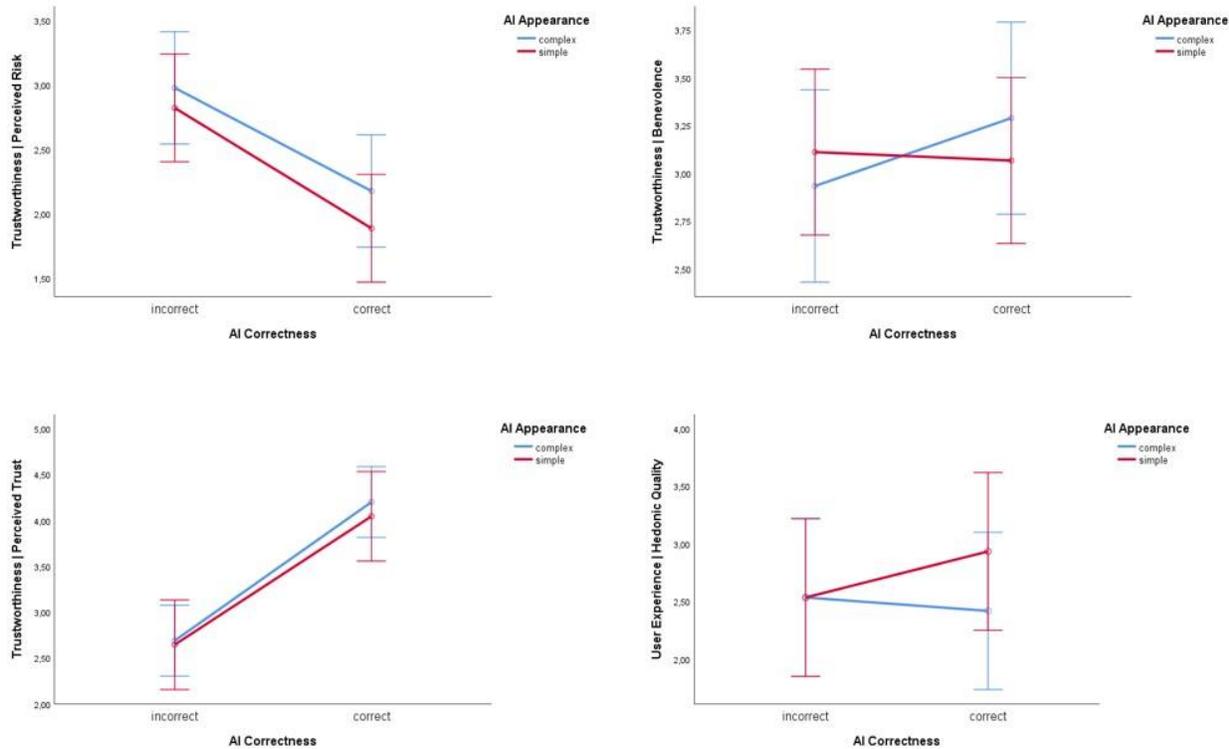

**Figure 10** shows the results of the trustworthiness ratings and the rating of the hedonic quality.

## Discussion

AI-applications are omnipresent, and AI as a term certainly implies far-reaching connotations for many users, not experts. HCI approaches would suggest evaluating the impact of different AI applications, appearances, and operations on human perception of and interaction with AIs and identifying the effects these manipulations would have on users. However, this approach, which is central to HCI principles, is challenging to realize due to the sheer complexity of many fields of applications of the real physical world. Another rising technology cluster, XR, accounts for control and systematic manipulations of complex interactions (Blascovich et al., 2002; Wienrich and Gramlich, 2020; Wienrich et al., 2020). Hence, XR provides much potential to increase the investigability of human-AI interactions and interfaces.

The present paper suggests and discusses new perspective on the XR-AI combination space and XR as a new testbed for human-AI interactions and interfaces to establish a valid and systematic investigation procedures for human-AI interfaces and interactions. Four sub-questions structured the article's first, theoretical, part (see Table 6). The second part presented two paradigmatic implementations of an XR testbed for human-AI interactions (see Table 7).

**Table 6** summarizes the results of the article's first part.

| Contribution | Research Question | Short Answer |
|---|---|---|
| A new perspective on the XR-AI combination space | How is human-AI interaction defined? | As Human-Computer Interaction focusing on the interaction between humans and Artificial Intelligences |
| | How to classify combined XR with AI? | XR-AI Continuum (see Figure 3) |
| A new testbed for human-AI interactions and interface | What can we learn from the challenges and XR solutions concerning the investigability of human-human interaction? | XR provides high ecological validity and high standardization for investigating social interactions between humans |
| | Which challenges and solutions arise for the systematic investigation of human-AI interaction and interfaces? | Challenges concerning interactive embodied AI:<br><br>artificial tool/device → artificial (social) counterpart<br><br>usage by skilled experts → diverse users usage<br><br>applications in specific domains → diverse (every day) domain applications<br><br>usage avoidance → incorrect usage with considerable consequences |
| | | XR as valid and systamtic testbed for HAI:<br><br>4 Potentials: Design Space, Accessibility, Versatility, Tangible Taining<br><br>→ Rapid Prototyping of HAI |

## Contribution 1: A new perspective on the XR-AI combination space

Lieberman motivated an integrated combination of AI and computer graphics (Lieberman, 1996). Aylett and Luck (Aylett and Luck, 2000) denoted, more specifically, combinations of AI and XR technologies as Intelligent Virtual Environments. In both views, AI is conceptualized as enabling technology to improve the computer graphics or virtual environments. The XR-AI continuum extends the XR-AI combination space by introducing XR as a new investigability of HAI. It includes two perspectives on XR-AI combinations: Does the combination target AI as an object of investigation, e.g., to learn about the effects of a prospective AI embodiment on users? Or does the combination use AI as an enabling technology for XR, e.g., to provide techniques for multimodal interfaces? Often, XR-AI combinations will serve both tasks to some specific degree which allows to map them to the XR-AI continuum.

The XR-AI continuum provides a frame to classify XR-AI combinations. Such a classification serves multiple scientific purposes: First, it allows to systematically evaluate and rate different approaches of XR-AI combinations and hence it helps to structure the overall field. Second, it hopefully leads to an identification of best practices for certain approaches depending on the general perspective taken, and hence, third, that these best practices provide useful guidance, e.g., for replicating or reevaluating work and results. XR-AI combinations can drastically vary in terms of the required development efforts and complexities (Latoschik and Blach, 2008; Fischbach et al., 2017). However, many approaches targeting the eXtended AI range of the XR-AI continuum seem – in principle – to be realizable either by favorable loose couplings of the AI components or to require a rather limited completeness or development effort of the AI part. Regarding the latter, XR technology allows to adopt lightweight AI-mockups based on the Eliza principle or on Wizard-of-Oz scenarios to various degrees and to extensively investigate the relationship between humans and AIs as well as important usability and user experience aspects before an AI is fully functional. Hence, these investigations can be performed without the necessity to first develop a full-blown AI system and, in addition, an AI development process can be integrated with a user-centered design process to continuously optimize the human-AI interface. These rapid prototyping prospects lay the cornerstone to render XR as an ideal testbed to research human-AI interaction and interfaces.

**Contribution 2: A new testbed for human-AI interactions and interface**

Results stemming from human-human interactions have shown that XR provides high ecological validity and high standardization for investigating social interactions between humans (Blascovich et al., 2002; Bombari et al., 2015; Pan and Hamilton, 2018). AI applications, becoming increasingly interactive and embodied, lead up to essential changes from an HCI perspective and challenges of a valid and systematic investigability. XR as a new testbed for human-AI interactions and interfaces provides four potentials (see also Table 3) discussed along with the results of the paradigmatic implementations (see Table 7):

1) **Design Space:** XR provides a powerful design space. Relatively simple and easy to develop XR environments brought insights into the impact of different social intelligence cues and appearances on a robot and recommender evaluation. Instead of test situations asking participants to imagine the AI interaction with different robots (recommender systems) or showing pictures or vignettes, participants interacted with the embodied AI increasing the results' validity. Furthermore, such simple XR applications can serve as a starting point for systematic variations such as other cues of social intelligence, other appearances, other embodiments of interactive AIs, and combinations of these factors. The effort-cost trade-off is more worthwhile than producing new systems for each variation. Finally, features becoming possible or essential in the future or which are inconceivable in the presence can be simulated. In turn, creative design solutions can be developed, keeping pace with the time and fast-changing requirements.

2) **Accessibility:** XR can capture direct insights of diverse user groups. The paradigmatic experiment investigated student participants. However, the setup can easily be transferred to a testing room in a real industrial or leisure environment and be replicated with users who will collaborate with robots or recommender systems in the future. In addition, distributed VR systems are getting more and more available. Participants from different regions with different profiles, various degrees of expertise, or other human-centered factors can be tested with the same standardized scenario. Thus, a XR testbed can contribute to comparable results revealing individual preferences and needs during a human-AI interaction. As we could see in experiment 1, women rated the robot and interaction differently from men, particularly concerning the need fulfillment. Moreover, interactions between variations of the AI or the human-AI interaction and various users can be investigated systematically.

3) **Versatility**: XR environments can simulate different domains or tasks such as industry or game. The current experiments simulated simple tasks in two settings. The tasks and the environments can easily be varied in XR. From the perspective of a human-centered design process, the question arises, which AI features are domain-specific and which general design guidelines can be deduced. By testing the same AI in different environments, the question can be answered. Further senseful fields of application can be identified – the same AI can be accepted by users in an industrial setting but not at home, for example. The task difficulty can also impact the evaluation of the AI. The presented results of experiment 1, for example, are valid for easy tasks where conversational abilities of the robot was rated positively. However, during more difficult tasks, conversation might be inappropriate due to distraction and consequences on the task performance. In reality, investigation of such impacts bears many risks while XR provides a safe testbed.

4) **Tangible Training:** XR can serve as a tangible training environment. The presented experiment did not include a training. However, the operational principles of such embodied AIs can be illustrated directly during the interaction. Users can learn step by step how to interact with AIs, what they can expect from AI applications, what abilities they have, and what limits AIs possess. In other words, users can get a realistic picture of the potentials and limits of AI applications. Again, operations becoming possible and essential in the future can already be simulated in XR. That possibility offers a user-centered design and training approaches keeping pace with the time and fast-changing requirements. Acceptance, expertise, and experience with AI systems can be developed proactively instead of retrospectively.

**Table 7** summarizes the results of the article's second part.

| Contribution | Research Question \| Experimental Approach | Paradigmatic Results |
|---|---|---|
| Experiment 1 simulated a robot-human interaction in an industrial context. | How does thesese of the robot's social intelligence influence the robot's perception and the evaluation of the interaction? 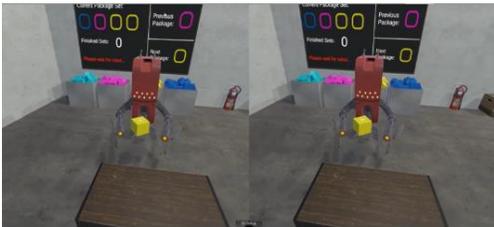 Manipulation of the conversational ability to vary the sense of social intelligence of a simulated robot. | The conversational robot was evaluated as more social, human-like, and positive than the non-conversational robot. The interaction with the conversational robot fulfilled more user motives than the non-conversational robot. Female users evaluated the robot and the interaction differently from male users. |
| Experiment 2 simulated a recommender system in a quiz game setting. | How does the sense of complexity influence the perception of a recommender system? | The simple appearance was evaluated as friendlier and less risky. |

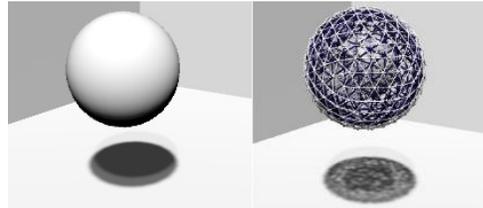

The interaction with the simple AI was more hedonic than the complex AI (when the AI answered 100% correctly).

Manipulation of the appearance of an embodied recommender system to vary the sense of complexity.

## Limitations

Although, we have many indicators stemming from investigating human-human interactions in XR (Blascovich, 2002; Bombari et al., 2015; Pan and Hamilton, 2018) and product testing in XR, simulations only resemble real human-AI interactions. Hence, we cannot be sure that the results stemming from XR interactions precisely predict human-AI interactions in real settings. For establishing XR as valid testbed, comparative studies testing real interaction settings with XR interactions would be necessary. Additionally, the rather uncomplicated variability of features might mislead researchers into engendering myriad results stemming from neglectable variation. The significance of the outcome would blow out of portion. Since XR simulations can be set up in each laboratory, results might increasingly be collected with student samples instead of real users of the corresponding AI application. Thus, it seems important to make XR testbeds attractive and available for practical stakeholders.

## Conclusion

AI covers a broad spectrum of computational problems and use cases. At the same time, AI applications are becoming increasingly interactive. Consequently, AIs will exhibit a certain form of embodiment to users at the human-AI interface. This embodiment is manifold and can, e.g., range from simple feedback systems, to text or graphical terminals, to humanoid agents or robots of various appearances. Simultaneously, human-AI interactions gain massively in importance for diverse user groups and in various fields of application. However, the outer appearance and behavior of an AI result in profound differences on how an AI is perceived by users, e.g., what they think about the AI's potential capabilities, competencies or even dangers, i.e., how user's attitude towards an AI are shaped. Moreover, many users or future users do have abstract ideas of what AI is, significantly increasing the embodiment effects and hence the importance of the specific embodiment of AI applications. These circumstances raise intricate questions of why and how humans interact or should interact with AIs.

From a human-computer interaction and human-centered design perspective, it is essential to investigate the acceptance of AI applications and the consequences of interacting with an AI. However, the sheer complexity of application fields and AI embodiments, in reality, face researchers with enormous challenges regarding valid and systematic investigations. This article proposes eXtended AI as a method for such a systematic investigation. We provided a theoretical treatment and model of Human-AI interaction based in the introduced XR-AI continuum as a framework for and a perspective of different approaches of XR-AI combinations. We motivate eXtended AI as a specific XR-AI combination capable of helping us to learn about the effects of prospective human-AI interfaces and to show *why* the combination of XR and AI fruitfully contributes to a valid and systematic investigation of human-AI interactions and interfaces.

The article also provided two exemplary experiments investigating aforementioned approach for two distinct AI-systems. The first experiment reveals an interesting gender effect in human-robot interaction, while the second experiment reveals an Eliza effect of a recommender system. These two examples of paradigmatic implementations of the proposed XR testbed for human-AI interactions and interfaces show *how* a valid and systematic investigation can be conducted. It additionally reports on two interesting findings of embodied human-AI interfaces supporting the proposed idea to use XR as a testbed to investigate human-AI interaction. From an HCI perspective, eXtended AI (left side of the XR-AI continuum) constitutes a variant of rapid prototyping for HAI. The clue is that users interact with a prestage or a simulation instead of the fully developed product. In sum, the article opens new perspectives on how XR benefits for systematic investigations are essential for an evidence-based human-centered AI design, the design and evaluation of explainable AIs and tangible training modules, and basic research of human-AI interfaces and interactions.

## Literature

Aragon, S. R. (2003). Creating social presence in online environments. *New Dir. Adult Contin. Educ.* 2003, 57–68. doi:10.1002/ace.119.

Aylett, R., and Luck, M. (2000). Applying Artificial Intelligence to Virtual Reality: Intelligent Virtual Environments. *Appl. Artif. Intell.* 14, 3–32.

Bailenson, J., Aharoni, E., Beall, A., Guadagno, R., Dimov, A., and Blascovich, J. (2004). Comparing behavioral and self-report measures of agents' social presence in immersive virtual environments. in *Proceedings of the 7th Annual International Workshop on PRESENCE*, 1864–1105. Available at: http://web.stanford.edu/~bailenso/papers/presconf.pdf [Accessed March 13, 2021].

Bär, N., Hoffmann, A., and Krems, J. (2011). "Entwicklung von Testmaterial zur experimentellen Untersuchung des Einflusses von Usability auf Online-Trust.," in *Reflexionen und Visionen der Mensch-Maschine-Interaktion – Aus der Vergangenheit lernen, Zukunft gestalten*, eds. S. Schmid, M. Elepfandt, J. Adenauer, and A. Lichtenstein, 627–631.

Bawden, D., and others (2008). Origins and concepts of digital literacy. *Digit. literacies Concepts, policies Pract.* 30, 17–32.

Biocca, F. (1999). The cyborg's dilemma: Progressive embodiment in virtual environments. *Hum. Factors Inf. Technol.* 113, 113–144.

Blascovich, J. (2002). "Social Influence within Immersive Virtual Environments," in *The social life of avatars* (Londond: Springer), 127–145. doi:10.1007/978-1-4471-0277-9_8.

Blascovich, J., Loomis, J., Beall, A. C., Swinth, K. R., Hoyt, C. L., and Bailenson, J. N. (2002). Immersive virtual environment technology as a methodological tool for social psychology. *Psychol. Inq.* 13, 103–124. doi:10.1207/S15327965PLI1302_01.

Bombari, D., Schmid Mast, M., Canadas, E., and Bachmann, M. (2015). Studying social interactions through immersive virtual environment technology: virtues, pitfalls, and future challenges.

*Front. Psychol.* 6. doi:10.3389/fpsyg.2015.00869.

Cameron, J. (1984). The Terminator.

Carolus, A., Binder, J. F., Muench, R., Schmidt, C., Schneider, F., and Buglass, S. L. (2019). Smartphones as digital companions: Characterizing the relationship between users and their phones. *journals.sagepub.com* 21, 914–938. doi:10.1177/1461444818817074.

Carolus, A., and Wienrich, C. (2019). How close do you feel to your devices? Visual assessment of emotional relationships with digital devices. *dl.gi.de*. doi:10.18420/muc2019-ws-652.

Carpenter, J. (1974). Dark Star.

Cavazza, M., Lugrin, J.-L., Pizzi, D., and Charles, F. (2007). Madame Bovary on the Holodeck: Immersive Interactive Storytelling. in *Proceedings of the 15th international conference on Multimedia* MULTIMEDIA '07. (New York, NY, USA: ACM), 651–660.

Clarke, A. C. (1962). Hazards of prophecy: The failure of imagination. *Profiles Futur.* 6, 1.

Congdon, S. P., and Schober, M. F. (2002). How examiners' discourse cues affect scores on intelligence tests. in (Kansas City, MO: 43th Annual Meeting of the Psychonomic Society). Available at: https://scholar.google.com/scholar_lookup?title=How+examiners%27+discourse+cues+affect+s cores+on+intelligence+test.&journal=Paper+Presented+at+the+43th+Annual+Meeting+of+the+ Psychonomic+Society&author=Congdon+S.+P.&author=and+Schober+M.+F.&publication_yea [Accessed March 13, 2021].

Cruz-Neira, C., Sandin, D. J., DeFanti, T. A., Kenyon, R. V, and Hart, J. C. (1992). The CAVE: audio visual experience automatic virtual environment. *Commun. ACM* 35, 64–72. doi:http://doi.acm.org/10.1145/129888.129892.

DiSessa, A. A. (2001). *Changing minds: Computers, learning, and literacy*. Mit Press.

Field, A. (2009). *Discovering Statistics Using SPSS, Thrid Edition*. SAGE Publications Available at: http://sutlib2.sut.ac.th/sut_contents/H124897.pdf [Accessed March 13, 2021].

Fischbach, M., Wiebusch, D., and Latoschik, M. E. (2017). Semantic Entity-Component State Management Techniques to Enhance Software Quality for Multimodal VR-Systems. *IEEE Trans. Vis. Comput. Graph.* 23, 1407–1416. doi:10.1109/TVCG.2017.2657098.

Fiske, S. T., Cuddy, A. J. C., Glick, P., and Xu, J. (2002). A model of (often mixed) stereotype content: Competence and warmth respectively follow from perceived status and competition. *J. Pers. Soc. Psychol.* 82, 878–902. doi:10.1037/0022-3514.82.6.878.

Hart, S. G., and Staveland, L. (1988). Development of NASA-TLX (Task Load Index): Results of empirical and theoretical research. in *Human mental workload* (P.A. Hancock and N. Meshkati (Eds.), Amsterdam: Elsevier), 139–183.

Hassenzahl, M., and Monk, A. (2010). The inference of perceived usability from beauty. *Human-Computer Interact.* 25, 235–260. doi:10.1080/07370024.2010.500139.

Hassenzahl, M., Wiklund-Engblom, A., Bengs, A., Hägglund, S., and Diefenbach, S. (2015). Experience-Oriented and Product-Oriented Evaluation: Psychological Need Fulfillment, Positive Affect, and Product Perception. *Int. J. Hum. Comput. Interact.* 31, 530–544. doi:10.1080/10447318.2015.1064664.

Hatfield, E., Cacioppo, J. T., and Rapson, R. L. (1992). Primitive emotional contagion. *Emot. Soc. Behav.* 14, 151–177. Available at: https://www.researchgate.net/publication/232480409 [Accessed March 13, 2021].

Ho, C. C., and MacDorman, K. F. (2010). Revisiting the uncanny valley theory: Developing and validating an alternative to the Godspeed indices. *Comput. Human Behav.* 26, 1508–1518. doi:10.1016/j.chb.2010.05.015.

Huta, V. (2016). "Eudaimonic and Hedonic Orientations: Theoretical Considerations and Research Findings," in *Handbook of eudaimonic well-being* (Springer), 215–231. doi:10.1007/978-3-319-42445-3_15.

Jonas K., Stroebe W., H. M. (2014). *Sozialpsychologie*. 6th ed. Springer-Verlag.

Kelley, P. G., Yang, Y., Heldreth, C., Moessner, C., Sedley, A., Kramm, A., et al. (2019). Happy and assured that life will be easy 10years from now: Perceptions of artificial intelligence in 8 countries∗†. *arXiv*. Available at: https://arxiv.org/abs/2001.00081 [Accessed March 13, 2021].

Kistler, F., Endrass, B., Damian, I., Dang, C. T., and André, E. (2012). Natural Interaction with Culturally Adaptive Virtual Characters. *Multimodal User Interfaces* 6, 39–47. doi:10.1007/s12193-011-0087-z.

Kubrick, S. (1968). 2001: A Space Odyssey.

Kuhlen, A. K., and Brennan, S. E. (2013). Language in dialogue: When confederates might be hazardous to your data. *Psychon. Bull. Rev.* 20, 54–72. doi:10.3758/s13423-012-0341-8.

Latoschik, M. E. (2005). A User Interface Framework for Multimodal {VR} Interactions. in *Proceedings of the IEEE seventh International Conference on Multimodal Interfaces, ICMI 2005* (Trento, Italy).

Latoschik, M. E. (2014). Smart Graphics/Intelligent Graphics. *Inform. Spektrum.*

Latoschik, M. E., and Blach, R. (2008). Semantic Modelling for Virtual Worlds -- A Novel Paradigm for Realtime Interactive Systems? in *Proceedings of the ACM VRST 2008* , 17–20.

Latoschik, M. E., Roth, D., Gall, D., Achenbach, J., Waltemate, T., and Botsch, M. (2017a). The effect of avatar realism in immersive social virtual realities. in *Proceedings of the 23rd ACM Symposium on Virtual Reality Software and Technology (VRST)*, 39:1-39:10. doi:10.1145/3139131.3139156.

Latoschik, M. E., Roth, D., Gall, D., Achenbach, J., Waltemate, T., and Botsch, M. (2017b). The Effect of Avatar Realism in Immersive Social Virtual Realities. in *23rd ACM Symposium on Virtual Reality Software and Technology (VRST)*, 39:1-39:10.

Latoschik, M. E., and Wachsmuth, I. (1998). *Exploiting distant pointing gestures for object selection in a virtual environment*. doi:10.1007/BFb0052999.

Lieberman, H. (1996). Intelligent graphics. *Commun. ACM* 39, 38–48. doi:10.1145/232014.232026.

Long, D., and Magerko, B. (2020). What is AI Literacy? Competencies and Design Considerations. in *Conference on Human Factors in Computing Systems - Proceedings* (Association for Computing Machinery), 1–16. doi:10.1145/3313831.3376727.

Mori, M., MacDorman, K. F., and Kageki, N. (2012). The uncanny valley [from the field]. *IEEE Robot. Autom. Mag.* 19, 98–100.

Nass, C., and Gong, L. (2000). Speech interfaces from an evolutionary perspective. *Commun. ACM* 43, 36–43. doi:10.1145/348941.348976.

Nomura, T., Suzuki, T., Kanda, T., and Kato, K. (2006). Measurement of negative attitudes toward robots. *Interact. Stud. Soc. Behav. Commun. Biol. Artif. Syst. Stud.* 7, 437–454. doi:10.1075/is.7.3.14nom.

Obaid, M., Damian, I., Kistler, F., Endrass, B., Wagner, J., and André, E. (2012). Cultural Behaviors of Virtual Agents in an Augmented Reality Environment. in *12th International Conference on Intelligent Virtual Agents (IVA 2012)* LNCS., eds. Y. I. Nakano, M. Neff, A. Paiva, and M. A. Walker (Springer), 412–418. doi:10.1007/978-3-642-33197-8_42.

Pan, X., and Hamilton, A. F. de C. (2018). Why and how to use virtual reality to study human social interaction: The challenges of exploring a new research landscape. *Br. J. Psychol.* 109, 395–417. doi:10.1111/bjop.12290.

Pham, D. T., and Gault, R. S. (1998). A comparison of rapid prototyping technologies. *Int. J. Mach. Tools Manuf.* 38, 1257–1287. doi:10.1016/S0890-6955(97)00137-5.

Ratan, R., Beyea, D., Li, B. J., and Graciano, L. (2019). Avatar characteristics induce users' behavioral conformity with small-to-medium effect sizes: a meta-analysis of the proteus effect. *Taylor Fr.* 23, 651–675. doi:10.1080/15213269.2019.1623698.

Razzouk, R., and Shute, V. (2012). What Is Design Thinking and Why Is It Important? *Rev. Educ. Res.* 82, 330–348. doi:10.3102/0034654312457429.

Reeves, B., and Nass, C. (1996a). *The media equation: How people treat computers, television, and new media like real people*. Cambridge university press Cambridge, UK.

Reeves, B., and Nass, C. (1996b). *The media equation: How people treat computers, television, and new media like real people and places.* Cambridge University Press Available at: https://psycnet.apa.org/record/1996-98923-000 [Accessed February 11, 2020].

Russell, S. J., and Norvig, P. (2020). *Artificial Intelligence: A Modern Approach*. 4th ed. Prentice Hall.

Schreier, J. (2012). Robot & Frank.

Scott, R. (1979). Alien.

Scott, R. (1982). Blade Runner.

Terry, L. R., and Garrison, A. D. R. (2007). Assessing Social Presence In Asynchronous Text-based Computer Conferencing J. *J. Distance Educ.* 14, 1–18. Available at: http://cade.athabascau.ca/vol14.2/rourke_et_al.html [Accessed March 13, 2021].

Topál, J., Gergely, G., Miklósi, Á., Erdőhegyi, Á., and Csibra, G. (2008). Infants' perseverative search errors are induced by pragmatic misinterpretation. *science.sciencemag.org* 321, 1831-1834. doi:10.1126/science.1113634.

Wardrip-Fruin, N. (2001). Three play effects – Eliza, Tale-Spin, and SimCity. *Digit. Humanit.*, 1–2. Available at: https://citeseerx.ist.psu.edu/viewdoc/download?doi=10.1.1.105.2025&rep=rep1&type=pdf [Accessed March 13, 2021].

Wienrich, C., Döllinger, N. I., and Hein, R. (2020). Mind the gap: A framework (BehaveFIT) guiding the use of immersive technologies in behavior change processes. *arXiv Prepr. arXiv2012.10912*. Available at: https://arxiv.org/abs/2012.10912 [Accessed March 13, 2021].

Wienrich, C., and Gramlich, J. (2020). appRaiseVR – An Evaluation Framework for Immersive Experiences. *i-com* 19, 103–121.

Wienrich, C., Gross, R., Kretschmer, F., and Müller-Plath, G. (2018a). Developing and Proving a Framework for Reaction Time Experiments in VR to Objectively Measure Social Interaction with Virtual Agents. in *IEEE VR*.

Wienrich, C., Schindler, K., Döllinger, N., and Kock, S. (2018b). Social Presence and Cooperation in Large-Scale Multi-User Virtual Reality-The Relevance of Social Interdependence for Location-Based Environments. in *IEEE Conference on Virtual Reality and 3D User Interfaces* (IEEE), 207–214. Available at: https://ieeexplore.ieee.org/abstract/document/8446575/ [Accessed July 26, 2020].

Yan, X., and Gu, P. (1996). A review of rapid prototyping technologies and systems. *CAD Comput. Aided Des.* 28, 307–318. doi:10.1016/0010-4485(95)00035-6.

Yee, N., and Bailenson, J. (2007). The proteus effect: The effect of transformed self-representation on behavior. *Hum. Commun. Res.* 33, 271–290. doi:10.1111/j.1468-2958.2007.00299.x.

Zhang, B., and Dafoe, A. (2019). Artificial Intelligence: American Attitudes and Trends. *SSRN Electron. J.* doi:10.2139/ssrn.3312874.

Zimmerer, C., Fischbach, M., and Latoschik, M. E. (2018a). Semantic Fusion for Natural Multimodal Interfaces using Concurrent Augmented Transition Networks. *Multimodal Technol. Interact.* 2, 81.

Zimmerer, C., Fischbach, M., and Latoschik, M. E. (2018b). Space Tentacles - Integrating Multimodal Input into a VR Adventure Game. in *Proceedings of the 25th IEEE Virtual Reality (VR) conference* (IEEE), 745–746.

**Conflict of Interest**

*The authors declare that the research was conducted in the absence of any commercial or financial relationships that could be construed as a potential conflict of interest.*

**Author Contribution**

The authors have contributed equally to this work.